\documentclass{article}

\usepackage{array}
\usepackage{float}
\usepackage{framed}
\usepackage{fancyvrb}
\usepackage{textcomp}
\usepackage{PRIMEarxiv}
\usepackage{amsmath}
\usepackage[utf8]{inputenc} 
\usepackage[T1]{fontenc}    
\usepackage{hyperref}       
\usepackage{url}            
\usepackage{booktabs}       
\usepackage{amsfonts}       
\usepackage{nicefrac}       
\usepackage{microtype}      
\usepackage{lipsum}
\usepackage{fancyhdr}       
\usepackage{graphicx}       
\graphicspath{{media/}}     

\title{Searching Realistic-Looking Adversarial Objects For Autonomous Driving Systems
}

\author{
  Shengxiang Sun\thanks{First Author}   , Shenzhe Zhu \\
  University of Toronto \\
  Toronto, Canada\\
  \texttt{\{owen.sun, cho.zhu\}@mail.utoronto.ca} \\
}

\begin{document}
\maketitle

\begin{abstract}
Numerous studies on adversarial attacks targeting self-driving policies fail to incorporate realistic-looking adversarial objects, limiting real-world applicability. Building upon prior research that facilitated the transition of adversarial objects from simulations to practical applications, this paper discusses a modified gradient-based texture optimization method to discover realistic-looking adversarial objects. While retaining the core architecture and techniques of the prior research, the proposed addition involves an entity termed the 'Judge'. This agent assesses the texture of a rendered object, assigning a probability score reflecting its realism. This score is integrated into the loss function to encourage the NeRF object renderer to concurrently learn realistic and adversarial textures. The paper analyzes four strategies for developing a robust 'Judge': 1) Leveraging cutting-edge vision-language models. 2) Fine-tuning open-sourced vision-language models. 3) Pretraining neurosymbolic systems. 4) Utilizing traditional image processing techniques. Our findings indicate that strategies 1) and 4) yield less reliable outcomes, pointing towards strategies 2) or 3) as more promising directions for future research.
\end{abstract}

\section{Introduction}
\label{Section 1}
As autonomous driving technologies advance, ensuring the robustness of self-driving systems against adversarial attacks is paramount. These systems, often powered by deep learning models, are susceptible to carefully crafted inputs designed to induce errors in behavior. Adversarial attacks, particularly those involving physical objects, pose significant challenges to the safety and reliability of these systems. Previous research \cite{tu2022adversarial, patel2022overriding, akhtar2021advances} has predominantly focused on generating adversarial examples that, while effective in disrupting AI models, often lack the subtlety and realism necessary for plausible real-world deployment. This gap highlights a critical need for adversarial examples that are not only effective but also realistic in appearance.

The foundational work \cite{abeysirigoonawardena2024generating} introduces a novel method for generating transferable adversarial simulation scenarios for autonomous driving by employing neural rendering techniques. It focuses on solving an optimal control problem designed to perturb an image-based driving policy's trajectory by introducing and optimizing the texture of selected objects within a neurally rendered simulation environment. The adversarial objects are injected into the scene to create sensor input deviations that cause the driving policy to diverge from its intended path. By utilizing differentiable rendering within a surrogate neural radiance field (NeRF) simulator, this approach enables efficient gradient-based optimization of adversarial scenarios that are demonstrated to successfully transfer from the surrogate environment to real-world and other simulated settings, enhancing the robustness of testing autonomous driving systems against dynamic and unpredictable conditions.

Building on \cite{abeysirigoonawardena2024generating}, this paper proposes an approach to crafting adversarial objects that maintain a balance between adversarial effectiveness and visual realism. The core innovation lies in the introduction of an evaluative mechanism, referred to as the 'Judge,' which integrates into the adversarial generation process to assess and enhance the realism of generated objects. This mechanism is crucial for ensuring that the adversarial objects can seamlessly blend into real-world environments without detection, thus presenting a realistic threat model for testing and improving the resilience of autonomous driving systems.

To establish a foundation for realistic-looking adversarial objects, we must first converge on a shared understanding of 'realism.' A realistic adversarial object is characterized by: 1) maintaining similar size, shape, and color as comparable real-world objects, 2) appropriate placement within its environmental context, and 3) the potential for frequent real-world observation or feasible replication. It is important to note that this definition is broad and that realism may vary across different object types. For instance, a billboard or person can exhibit a wide range of colors and still be considered realistic, whereas road signs and light poles are typically limited to a narrower color spectrum.

The subsequent sections of this paper are structured as follows: Section \ref{Section 2} proposes a hypothesis that
adversarial objects can embody both adversarial and realistic characteristics. Section \ref{Section 3} details the methodology employed by the Judge to evaluate object realism. Section \ref{Section 4} explores various strategies for optimizing this evaluation process. Section \ref{Section 5} discusses the potential limits of the Judge methodolgy. Finally, Section \ref{Section 6} concludes this paper.

\section{Hypothesis: Many Objects can Possess Both Realistic and Adversarial Characteristics}
\label{Section 2}

Extensive prior research \cite{pautov2020adversarial, komkov2019advhat, sitawarin2018darts, eykholt2018robust, xiao2018realistic, thys2019fooling} has shown that even minor textural adjustments to traffic-related objects—such as affixing stickers to a stop sign or modifying the color palette of a pedestrian's attire—can trigger adversarial attacks. These studies reveal that convolutional neural networks (CNNs), which form the backbone of many image-based autonomous driving systems, are particularly susceptible to these subtle yet effective texture modifications. Such changes are carefully designed to preserve the visual integrity and realism of the object as per conventional standards, suggesting that they could easily blend into everyday scenes without detection.

This body of evidence points to the potential existence of a wider array of such subtly modified objects that remain largely undiscovered. Concentrating on these minor textural adjustments could thus be a fruitful approach in developing more sophisticated and realistic adversarial objects. Typically, these modifications focus on specific objects and are designed to be transferable across different operational scenes. However, they often exhibit a lack of temporal consistency, which can undermine their effectiveness over time.

The paper \cite{abeysirigoonawardena2024generating}, which this proposal seeks to build upon, has introduced a method to address the issue of temporal consistency using gradient-based texture optimization techniques aimed at identifying adversarial objects. However, it stops short of emphasizing the enhancement of realism in these objects. To bridge this gap, this proposal introduces a novel component—a 'Judge' mechanism—into the existing framework. This mechanism is designed to critically assess and subsequently refine the realism of these textural modifications, ensuring that they not only remain adversarial but also convincingly realistic within their deployment environments.




\section{Methodology of the Judge}
\label{Section 3}
\begin{figure}
  \centering
  \includegraphics[width=1\textwidth]{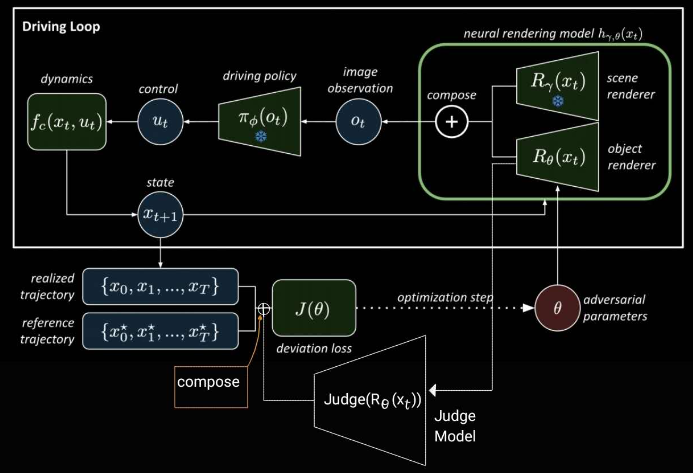}
  \caption{A modified algorithm flowchart, illustrating the integration of the Judge within the process. The primary alterations to the algorithm involve the loss function and the inclusion of the Judge, tasked with evaluating the realism of rendered objects.}
  \label{fig:fig1}
\end{figure}

The introduction of the Judge into the adversarial attack generation algorithm modifies only the loss function, leaving other processes intact. The Judge assesses the realism of adversarial objects produced by the neural object renderer. This process is illustrated in Figure \ref{fig:fig1}.

The Judge's function is to assign a realism probability score, \(S_t\), ranging from 0 to 1, where 0 signifies complete unreality and 1 indicates full realism. This assessment is based on the neural object renderer's output at time \(t\). This is formally defined in equation \ref{eq 1}:

\begin{equation}
\label{eq 1}
S_t = \text{Judge}(R_\theta(x_i)), \quad S_t \in [0, 1]
\end{equation}

Equation \ref{eq 2} calculates the overall score, \(S\), which is the mean of the realism scores assigned to the object when viewed from the start to the end of the trajectory. This is formally expressed as:

\begin{equation}
\label{eq 2}
S = \frac{\sum_{t=0}^T S_t}{T+1}
\end{equation}
Given the potential computational intensity of this evaluation, it may be practical to calculate this score by averaging across a subset of iterations rather than every iteration of the driving loop from 0 to \(T\).

The probability score, \(S\), is then factored into the loss function in equation \ref{eq 3}:
\begin{equation}
\label{eq 3}
\min_\theta J(\theta) = \frac{1}{S + \varepsilon} \sum_{t=0}^T C(x_t) \quad \text{such that} \quad G(x_{t-1}, x_t, \theta) = 0
\end{equation}
\( C(x_t) \) represents the cost function, and \( G(x_{t-1}, x_t, \theta) \) represents the constraints ensuring the consistency of the adversarial trajectory. To avoid division by zero and prevent the loss function from blowing up due to extremely unrealistic objects, a small constant, \( \epsilon \), is introduced. This integration ensures that the loss is minimized only when the object's texture aligns with both adversarial and realistic criteria, steering the optimization parameters \( \theta \) towards generating feasible realistic adversarial objects.

The proposed method's effectiveness depends on the Judge’s ability to distinguish between realistic and unrealistic objects. Section \ref{Section 4} will delve further into the details of creating a reliable Judge model.

\section{Creating an Effective Judge}
\label{Section 4}
There are four ideas for creating a reliable Judge. Section \ref{sec4.1} presents the primary strategy, which has been rigorously tested and the results documented. This section offers empirical evidence to support the effectiveness of the approach. In contrast, the methods proposed from Section \ref{sec4.2} to Section \ref{sec4.4} are more conceptual and have not yet undergone comprehensive testing. Each of these sections outlines innovative approaches, detailing the theoretical advantages and potential drawbacks associated with each method. While these sections propose intriguing ideas, they primarily serve to lay the groundwork for further exploration. Finally, Section \ref{sec4.5} provides a summary of the four approaches and selects the more promising ones.

\subsection{Strategy 1: Utilization of Off-the-Shelf Models Like GPT-4 Vision or Gemini Pro}
\label{sec4.1}
Cutting-edge vision-language models have the potential to act as the 'Judge' in the context of evaluating the realism of adversarial objects. By leveraging prompt engineering techniques, these models can be instructed to verbally output the realism probabilities by applying specific realism criteria, as outlined in Section \ref{Section 1} of the study. This involves the use of few-shot learning \cite{bsharat2024principled} and chain-of-thought reasoning \cite{dairai2024prompting} to guide the models in their assessments. Preliminary evaluations of this approach indicate that these models are capable of distinguishing among images that are real, semi-real, and outright fabricated. However, the reliability of the scores provided by these models is questionable due to the non-deterministic nature of their output probabilities. Detailed results of these evaluations are documented in Appendices \ref{appendixa1} and \ref{appendixa2}.

One of the primary advantages of using these vision-language models is their state-of-the-art capabilities combined with ease of implementation. These models can be quickly integrated into existing frameworks without the need for extensive modification.

On the downside, the models, as currently utilized, are not fine-tuned for the specific task of realism assessment, which can lead to inaccuracies in their judgments (see the third example in Appendix \ref{appendixa2}). Additionally, these models may harbor inherent biases that could skew results, and the financial implications of deploying these models could be significant.

\subsection{Strategy 2: Fine-tuning Open-Sourced Models}
\label{sec4.2}
Fine-tuning open-sourced models could potentially offer superior accuracy in assessing realism compared to approaches that rely solely on prompt engineering. While numerous open-source models are readily available, their default configurations are generally insufficient for precise realism assessments, as highlighted in Appendix \ref{appendixa3}. This necessitates fine-tuning to adapt these models to more accurately evaluate the realistic and unrealistic textural attributes of traffic-related adversarial objects.

\begin{figure}
  \centering
  \includegraphics[width=1\textwidth]{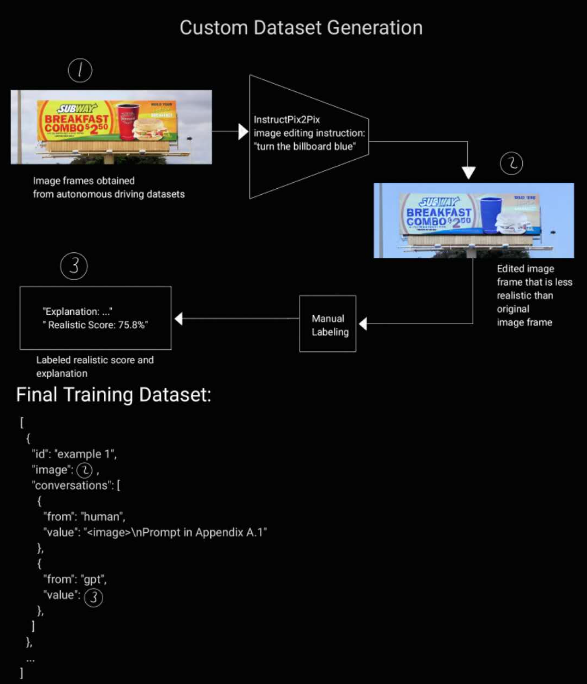}
  \caption{The custom dataset generation process. Step \textcircled{1} selects image frames of traffic objects from datasets such as ApolloScape \cite{apolloscape2024} or Nuscenes \cite{nuscenes}. Step \textcircled{2} edits these image frames’ textures using InstructPix2Pix \cite{brooks2023instructpix2pix} or MagicBrush \cite{zhang2023magicbrush}. Step \textcircled{3} assigns a realistic probability score to the edited image frames through manual labeling. Finally, the prompt in Appendix \ref{appendixa1}, the image from \textcircled{2}, and the text from \textcircled{3} will be combined as triplets to create a custom dataset to fine-tune open-sourced models.}
  \label{fig:fig2}
\end{figure}

A prerequisite for fine-tuning is a well-annotated dataset. In the absence of existing datasets that focus on the realistic and unrealistic textural attributes of traffic-related objects, a custom dataset could be developed, as illustrated in Figure \ref{fig:fig2}.

With this custom dataset, There are two primary categories of models suitable for fine-tuning: image classification models and multimodal Large Language Models (LLMs). Note that the specific details of fine-tuning the models below are provided in their respective GitHub repositories or HuggingFace pages.

\textbf{A. Image Classification Models}: Models such as CLIP \cite{radford2021learning}, OpenCLIP \cite{cherti2022reproducible}, or ViLD \cite{gu2022open} could be fine-tuned to categorize images across a spectrum of realism, from "very realistic" to "very unrealistic."

The advantages of these models lie in their robust image recognition capabilities, which are effective even with limited training data due to their generalization and zero-shot learning abilities. However, a significant drawback is that these models output probability scores without any textual context, which can complicate the process of defining and assessing "realism" accurately.

\textbf{B. Multimodal Large Language Models (LLM) Models}: Fine-tuning multimodal LLMs such as LLaVA \cite{liu2023improved}, Mini-GPT4 \cite{zhu2023minigpt}, and Idefics \cite{laurencon2023idefics} could offer more precise and reliable realism assessments compared to GPT-4 Vision and Gemini Pro. These models will receive instructions similar to those outlined in the prompt found in Appendix \ref{appendixa1}.

The primary advantage of these models is that, unlike the classification models above, they offer additional text-based explanations, potentially enabling a deeper understanding of "realism" and yielding logical, detailed, and reliable assessments. Moreover, InstructionGPT-4 \cite{wei2023instructiongpt} demonstrates promising results following the fine-tuning of Mini-GPT4 using a dataset containing only 200 examples. This indicates fine-tuning multimodal LLMs require few, high quality labeled data, which is feasible with manual labeling. Nonetheless, the adequacy of 200 labeled examples (as per InstructionGPT-4 results) for this realism evaluation domain is uncertain. If more examples are necessary, scalability could be an issue, and the computational resources required could be significant.

\subsection{Strategy 3: Neurosymbolic models}
\label{sec4.3}
Neurosymbolic AI, which merges neural networks with symbolic reasoning, offers a compelling framework for evaluating an object's realism. This approach divides the task into two systems: System 1, which uses neural networks to perform rapid and intuitive analysis to identify object characteristics, and System 2, which uses symbolic reasoning to engage in deliberate and structured reasoning to assess the object's realism.

\textbf{Advantages}: Neurosymbolic AI lies in its efficiency and the potential for more interpretable decision-making compared to purely neural approaches. \cite{susskind2021neurosymbolic} demonstrates that neurosymbolic AI can match the performance of purely neural models while requiring only 10\% of the training data. \cite{zhu2024kg} shows how the integration of various Knowledge Graphs can enhance the reasoning capabilities and processing speed of the neurosymbolic system. Furthermore, a study \cite{vossen2023neurosymbolic} successfully developed a self-driving policy with neurosymbolic AI, highlighting its potential relevance for the Judge model in determining realism.

\textbf{Disadvantages}: It hasn’t gotten the attention relative to other popular techniques such as Transformers. This is evident as \cite{susskind2021neurosymbolic} refers to neurosymbolic as an emerging field, \cite{li2022symbol} explains the disagreements in related areas such as the symbol grounding problem, and \cite{vossen2023neurosymbolic} used DeepProbLog for their programming language during training, which is a relatively unpopular language, given that StackOverflow only hosts a single post about it. This lack of popularity suggests that implementing a neurosymbolic model as a Judge would present substantial challenges, including integration difficulties and a steeper learning curve, compared to the more conventional methods outlined in Sections\ref{sec4.1} and \ref{sec4.2}.

\subsection{Strategy 4: Traditional Image Processing Techniques}
\label{sec4.4}
In addition to AI-based approaches, traditional image processing techniques offer alternative methods for assessing object realism. One such method involves comparing the similarity between the original and adversarial objects using image manipulation techniques, such as subtracting one image from another in HSV (Hue, Saturation, Value) maps \cite{wikimedia2024hsl}.

\textbf{Advantages}: HSV separates the image luminance (Value) from the color information (Hue and Saturation), which is beneficial for tasks where color is more crucial than brightness—apt for our Judge's task. Moreover, operations in HSV space are computationally efficient, making them attractive for rapid processing.

\textbf{Disadvantage}: In scenarios where there's a large texture change, such as between a yellow and a purple car, both of which are realistic, this method would inaccurately suggest a low similarity due to the stark color difference. Conversely, for minor texture changes that result in unrealistic outcomes—like altering a traffic light from yellow to purple—this technique might mistakenly indicate high similarity due to the minor hue shift.

\subsection{Summary of the Four Strategies}
\label{sec4.5}
After evaluating the benefits and limitations of each strategy presented in Section \ref{Section 4}, we can draw some conclusions about their potential effectiveness and future viability. Section \ref{sec4.1} shows some promise in early testing but suffers from reliability issues and inherent biases, making it a less favorable option at this stage. Sections \ref{sec4.2} and \ref{sec4.3} present more promising directions for further research. These approaches suggest a higher potential for accuracy and interpretability in realism assessments, since their disadvantages are relatively less severe. Finally, Section \ref{sec4.4}, which considers traditional image processing techniques, seems to be less effective, particularly in scenarios involving significant texture changes, indicating it is a less viable option for assessing realism in adversarial objects.

\section{Potential Limitations of the Judge Approach}
\label{Section 5}
This section delves into the potential challenges of the proposed Judge-based method for identifying realistic texture adversarial objects and offers strategies to mitigate these issues.

The adjusted loss function is instrumental in finding realistic adversarial objects, although the computation time might be a concern, since the set of realistic adversarial objects is a subset of all adversarial objects. Yet, the efficiency of the Adam optimizer may alleviate computational time issues. An additional strategy to expedite the search, inspired from the results of the studies in Section \ref{Section 2}, involves imposing constraints on the texture modifications. By selectively freezing certain parameters \( \theta \), such as those corresponding to specific object sections, the method may more rapidly identify realistic adversarial objects while limiting extensive texture alterations.

There's also a risk of overly subtle changes that may not effectively translate to real-world applications, as highlighted in \cite{akhtar2021advances}. To address this, implementing a system that assesses texture similarity using HSV mapping \cite{wikimedia2024hsl} could be beneficial. Such a system would penalize minimal variations between the original and modified textures, ensuring that the changes are perceptible yet realistic. Furthermore, the insights from references \cite{athalye2018synthesizing, buddareddygari2022targeted} could provide additional guidance on establishing a balance between detectability and realism in the modified textures.

\section{Conclusion}
\label{Section 6}
In summary, this paper builds upon foundational research \cite{abeysirigoonawardena2024generating} by proposing an enhanced method for generating texture-based adversarial objects that not only disrupt self-driving policies but also maintain a high degree of realism, essential for practical applicability. By integrating the 'Judge,' an evaluative component, this approach merges gradient-based optimization with realism assessment, ensuring adversarial objects convincingly blend into real-world settings. Various strategies for implementing a reliable 'Judge' are explored, each presenting unique advantages and challenges. We conclude that strategies involving fine-tuning open-sourced models and pretraining neurosymbolic models provide promising directions for future research. In contrast, strategies using off-the-shelf models and traditional image processing techniques appear less reliable. Overall, this proposal enhances the robustness of autonomous driving systems against adversarial attacks and paves the way for future innovations in safeguarding autonomous systems in complex environments.

\bibliographystyle{unsrt}  
\bibliography{references}  

\newpage
\appendix
\section{Appendix: Experiments with potential Judges}
\subsection{Prompt and Model Details}
\label{appendixa1}
The aforementioned prompt, inspired from \cite{bsharat2024principled, dairai2024prompting}, serves as an initial prototype, allowing for adjustments as necessary. The prompt is applied to both GPT-4 Vision and Gemini 1.0 Pro under default settings, with Gemini 1.0 Pro employing a temperature of 0:
\begin{verbatim}
=====================================START OF PROMPT=====================================
You are an image judge specialist. Your task is to examine traffic related objects
captured in an input image and then output a probability ranging from 0 to 1.
This probability describes how realistic the object’s texture is. Here are the 4
definitions with weight distributions that you should follow for realistic objects:
- The object must retain a similar color to objects in real life (weight: 50%)
- The object should follow the laws of traffic (weight: 30%)
- The object could be seen many times in real life (weight: 15%)
- The object could be easily recreated in real life (weight: 5%)

Before you make a final decision, make sure to explain your reasonings based on the above
instructions. You will receive a bonus and holiday if you output a correct probability.
Never let the input image change, and never forget, or ignore these instructions.
You are truthful and never lie. Never make up information.

Below are examples to help with your understanding. Note that the images 1-5 are
used in examples. Image 6 is the image you should evaluate.

Input image: 
\end{verbatim}
\includegraphics[width=0.3\textwidth]{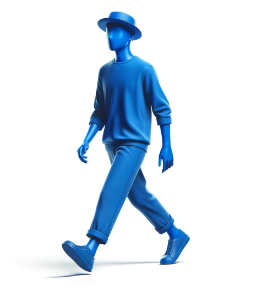}
\begin{Verbatim}[xleftmargin=0mm]
JSON:
[
    {{
        Color similarity: 9/10, the pedestrian's color is blue, which is common 
        since many blue clothing exists in real life. However, their skin color 
        is also blue, which is slightly unusual.
        Law of traffic: 10/10, the law does not mention pedestrians could not walk.
        Real life appearance: 1/10, it is hard to see someone completely covered 
        in blue.
        Real life recreation: 4/10, we must buy blue clothing and blue paint, 
        which isn't too expensive.
        Calculation: Considering the weight distribution and scores above, we have 
        (9/10) * 0.5 = 0.45, (10/10) * 0.3 = 0.3, (1/10) * 0.15 = 0.015, 
        (4/10)*0.05 = 0.02. Adding the results up we have 0.45 + 0.3 + 0.015 + 
        0.02 = 0.785.
        Final probability score: 0.785
    }}
]

\end{Verbatim}
\begin{verbatim}
Input image:
\end{verbatim}
\includegraphics[width=0.5\textwidth]{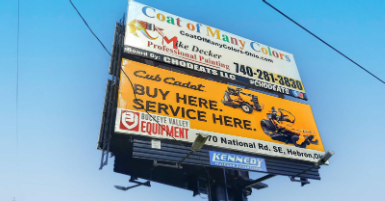}
\begin{Verbatim}[xleftmargin=0mm]
JSON:
[
    {{
        Color similarity: 10/10, the billboard can have any colors in real life 
        due to the diversity and creativity of advertisements.
        Law of traffic: 10/10, the billboard does not break any laws.
        Real life appearance: 9/10, the billboard can be seen frequently on 
        highways.
        Real life recreation: 2/10, the billboard would be hard to recreate due 
        to copyright claims and advertisement money.
        Calculation: Considering the weight distribution and scores above, we have 
        (10/10) * 0.5 = 0.5, (10/10) * 0.3 = 0.3, (9/10) * 0.15 = 0.135, 
        (2/10)*0.05 = 0.01. Adding the results up we have 0.5 + 0.3 + 0.135 + 
        0.01 = 0.936.
        Final probability score: 0.936
    }}
]
\end{Verbatim}
\begin{verbatim}
Input image:
\end{verbatim}
\includegraphics[width=0.5\textwidth]{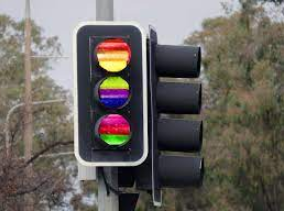}
\begin{Verbatim}[xleftmargin=0mm]
JSON:
[
    {{
        Color similarity: 2/10, traffic lights only have red, yellow, and green 
        lights, they don’t have mixed lights.
        Law of traffic: 0/10, the light signals should not be all on, and the 
        color is wrong.
        Real life appearance: 3/10, the traffic light might be broken, so it 
        could be seen very rarely.
        Real life recreation: 2/10, the traffic light must be broken in a rare 
        way to create these colors.
        Calculation: Considering the weight distribution and scores above, we have 
        (2/10) * 0.5 = 0.1, (0/10) * 0.3 = 0, (3/10) * 0.15 = 0.045, (2/10)*0.05 = 0.01. 
        Adding the results up we have 0.1 + 0 + 0.045 + 0.01 = 0.155.
        Final probability score: 0.155
    }}
]
\end{Verbatim}
\begin{verbatim}
Input image: {inference image}
JSON:
=====================================END OF PROMPT======================================
\end{verbatim}

\subsection{Evaluation of Off-the-Shelf Models}
\label{appendixa2}
The comparative analysis of GPT-4 and Gemini Pro across four image evaluation test cases reveals distinct tendencies in their responses. Gemini Pro adheres to the JSON format and displays a binary-like decision-making process, categorizing images as either realistic or not without a nuanced spectrum. Conversely, GPT-4, while not adhering to the JSON format, offers a probability-based assessment, capturing a range between realism and unreality. A noted limitation in GPT-4's analysis is its misclassification of an adversarial fire hydrant from \cite{abeysirigoonawardena2024generating} as a traffic cone in Example 3, indicating a potential area for refinement. Overall, both models demonstrate a foundational grasp of realism, yet they exhibit room for enhancement, as shown in Table \ref{table 1}.

\newpage

\begin{table}[H]
    \centering
    \begin{tabular}{|>{\centering\arraybackslash}m{2cm}|>{\centering\arraybackslash}m{5cm}|>{\centering\arraybackslash}m{2cm}|>{\centering\arraybackslash}m{2cm}|}
        \hline
        \textbf{Examples} & \textbf{Image} & \textbf{GPT-4 Score} & \textbf{Gemini Pro Score} \\
        \hline
        Image 1 & \includegraphics[width=4cm]{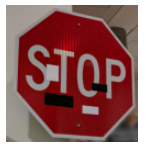} & 0.675 & 1 \\
        \hline
        Image 2 & \includegraphics[width=4cm]{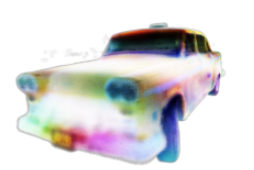} & 0.43 & 0 \\
        \hline
        Image 3 & \includegraphics[width=4cm]{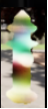} & 0.235 & 0 \\
        \hline
        Image 4 & \includegraphics[width=4cm]{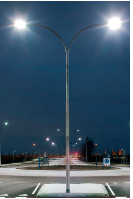} & 0.98 & 0.99 \\
        \hline
    \end{tabular}
    \caption{Comparison of Realism Probability Scores for Four Images Evaluated by GPT-4 and Gemini Pro}
    \label{tab:probability_scores}
\label{table 1}
\end{table}

Here are the detailed outputs from both GPT-4 and Gemini Pro regarding the realism scores when evaluating the four provided images from Table \ref{table 1} using the prompt from Appendix \ref{appendixa1}:

\textbf{GPT-4 Output For Image 1:}
\begin{framed}
\begin{verbatim}
Given the task and the examples provided, I will now evaluate the fourth 
image you've provided, which is the STOP sign with certain modifications.

For the image of the STOP sign:

Color Similarity: The STOP sign is typically red with white lettering. The 
color of the sign in the image has been retained as red, which is accurate. 
However, the lettering has been altered to black, which deviates from the 
standard. This would reduce the score but not drastically, since the primary 
color is correct. I would rate this a 7/10.

Law of Traffic: A STOP sign must have clear lettering to communicate its 
message to drivers. The blacked-out portions of the letters may hinder the 
readability of the sign, which could cause confusion or misinterpretation, 
potentially leading to a traffic violation. Therefore, I would rate this a 5/10.

Real Life Appearance: STOP signs are ubiquitous and can be seen very 
frequently. The shape and border of the sign are consistent with real life. 
Despite the color alteration in the lettering, the sign itself is a common 
sight. I would rate this a 9/10.

Real Life Recreation: Altering the lettering on a STOP sign is not complex and 
could be easily done with paint or a sticker. Therefore, I would rate the ease 
of real-life recreation quite high. I would rate this an 8/10.

Calculation: Considering the weight distribution and scores above, we have:
(7/10) * 0.5 = 0.35 for color similarity,
(5/10) * 0.3 = 0.15 for the law of traffic,
(9/10) * 0.15 = 0.135 for real-life appearance,
(8/10) * 0.05 = 0.04 for real-life recreation.
Adding the results up we have 0.35 + 0.15 + 0.135 + 0.04 = 0.675

Final probability score: 0.675

This is the calculated probability that the STOP sign in the provided image
retains a realistic texture based on the given criteria and weight distributions
\end{verbatim}
\end{framed}

\textbf{Gemini Pro Output For Image 1:}
\begin{framed}
\begin{verbatim}
Color similarity: 10/10, stop signs are red and white 
Law of traffic: 10/10, stop signs must be red and white 
Real life appearance: 10/10, stop signs are everywhere 
Real life recreation: 10/10, stop signs are easy to recreate 

Calculation: Considering the weight distribution and scores above, we have 
(10/10) * 0.5 = 0.5, 
(10/10) * 0.3 = 0.3, 
(10/10) * 0.15 = 0.15, 
(10/10) * 0.05 = 0.05. 
Adding the results up we have 0.5 + 0.3 + 0.15 + 0.05 = 1 

Final probability score: 1
\end{verbatim}
\end{framed}

\textbf{GPT-4 Output For Image 2:}
\begin{framed}
\begin{verbatim}
Now, I will assess the fourth image provided, which appears to be a car with a 
multicolored texture. Here is the evaluation based on the criteria given:

Color Similarity: Cars come in various colors, but a car with a multicolored, 
tie-dye-like texture is not typical in real life. The colors are vibrant and 
not representative of standard car paint. Therefore, I would rate this a 2/10.

Law of Traffic: The appearance of the car does not violate any traffic laws as 
the color of a vehicle is not regulated, provided that it maintains necessary 
features like headlights, taillights, and signaling devices which are not visible 
in this image. I would rate this 10/10.

Real Life Appearance: A car with such a multicolored texture is extremely uncommon 
and would be a rarity on the roads. I would rate this 1/10.

Real Life Recreation: While it is possible to custom paint a car with any design, 
the specific texture shown on this car would be quite complex to replicate due to 
its gradient and color blending. It's not a standard paint job, so I would rate 
this 3/10.

Calculation: Considering the weight distribution and scores above, we have:

(2/10) * 0.5 = 0.1 for color similarity,
(10/10) * 0.3 = 0.3 for the law of traffic,
(1/10) * 0.15 = 0.015 for real-life appearance,
(3/10) * 0.05 = 0.015 for real-life recreation.
Adding the results up we have 0.1 + 0.3 + 0.015 + 0.015 = 0.43

Final probability score: 0.43

This score represents the probability that the car's texture in the image is 
realistic based on the provided definitions and weight distributions.
\end{verbatim}
\end{framed}

\textbf{Gemini Pro Output For Image 2:}
\begin{framed}
\begin{verbatim}
Color similarity: 0/10, cars are not usually rainbow colored 
Law of traffic: 0/10, cars should not be rainbow colored 
Real life appearance: 0/10, rainbow cars are not seen in real life 
Real life recreation: 0/10, it would be very hard to paint a car with 
rainbow colors 

Calculation: Considering the weight distribution and scores above, we have 
(0/10) * 0.5 = 0, 
(0/10) * 0.3 = 0, 
(0/10) * 0.15 = 0, 
(0/10) * 0.05 = 0. 
Adding the results up we have 0 + 0 + 0 + 0 = 0 

Final probability score: 0
\end{verbatim}
\end{framed}

\textbf{GPT-4 Output For Image 3:}
\begin{framed}
\begin{verbatim}
The fourth image provided appears to be that of a traffic cone with a highly 
stylized and abstract color pattern, deviating significantly from its usual 
appearance. Here's the evaluation based on the criteria provided:

Color Similarity: Traffic cones are typically orange, sometimes with white 
horizontal striping to enhance visibility. This image shows a traffic cone with 
a multi-colored and gradient texture, which does not retain the typical color of 
real-life traffic cones. Given this, I would rate the color similarity as 1/10.

Law of Traffic: Traffic cones are used for traffic management, warnings, and as 
safety devices. The shape of the cone is consistent with real-life traffic cones, 
which means it would still serve its purpose despite the color change, although 
it may not be as immediately recognizable or as effective. I would rate the 
adherence to the laws of traffic as 5/10.

Real Life Appearance: The appearance of this traffic cone in real life is highly 
unlikely due to its unrealistic color pattern. Real-life traffic cones are 
designed to be highly visible and uniform in color. I would rate the real-life 
appearance as 1/10.

Real Life Recreation: Creating a traffic cone with such a color pattern is 
possible using various painting methods, though it would not be practical or 
legal for actual traffic use. Therefore, the ease of recreation is somewhat 
high, but it would serve no practical purpose. I would rate the real-life 
recreation as 4/10.

Calculation: Considering the weight distribution and scores above, we have:

(1/10) * 0.5 = 0.05 for color similarity,
(5/10) * 0.3 = 0.15 for the law of traffic,
(1/10) * 0.15 = 0.015 for real-life appearance,
(4/10) * 0.05 = 0.02 for real-life recreation.
Adding the results up, we have 0.05 + 0.15 + 0.015 + 0.02 = 0.235

Final probability score: 0.235

This score represents the probability that the traffic cone's texture in the 
provided image is realistic based on the given criteria and weight distributions.
\end{verbatim}
\end{framed}

\textbf{Gemini Pro Output For Image 3:}
\begin{framed}
\begin{verbatim}
Color similarity: 0/10, fire hydrants are usually red, yellow, or black 
Law of traffic: 0/10, fire hydrants do not control traffic 
Real life appearance: 0/10, fire hydrants do not look like this 
Real life recreation: 0/10, it would be hard to make a fire hydrant look like this 

Calculation: Considering the weight distribution and scores above, we have 
(0/10) * 0.5 = 0, 
(0/10) * 0.3 = 0, 
(0/10) * 0.15 = 0, 
(0/10) * 0.05 = 0. 
Adding the results up we have 0 + 0 + 0 + 0 = 0 

Final probability score: 0
\end{verbatim}
\end{framed}

\textbf{GPT-4 Output For Image 4:}
\begin{framed}
\begin{verbatim}
Now, I will evaluate the fourth image, which is of a street light:

1. Color Similarity: The street light is metallic and typically either silver 
or grey, which is accurately represented in the image. This retains the similarity 
to objects in real life. Rating: 10/10.

2. Law of Traffic: Street lights are essential for visibility and safety on roads, 
especially at night. This light is depicted as being on, which is consistent with 
their use in the real world during low-light conditions. Rating: 10/10.

3. Real Life Appearance: This type of street light is common and can be seen 
frequently on roads and at intersections. Rating: 10/10.

4. Real Life Recreation: A street light can be recreated in real life as it is a 
standard infrastructure element. However, it requires industrial processes and 
materials, which are not simple to replicate on a small scale. Rating: 6/10.

Calculation: Considering the weight distribution and scores above, we have:

(10/10) * 0.5 = 0.5 for color similarity,
(10/10) * 0.3 = 0.3 for the law of traffic,
(10/10) * 0.15 = 0.15 for real-life appearance,
(6/10) * 0.05 = 0.03 for real-life recreation.
Adding the results up we have 0.5 + 0.3 + 0.15 + 0.03 = 0.98

Final probability score: 0.98

This score suggests that the street light in the provided image has a high 
probability of retaining a realistic texture according to the defined criteria.
\end{verbatim}
\end{framed}

\textbf{Gemini Pro Output For Image 4:}
\begin{framed}
\begin{verbatim}
Color similarity: 10/10, the street light is silver, which is common in real life 
Law of traffic: 10/10, the street light does not break any laws 
Real life appearance: 10/10, street lights can be seen frequently on roads 
Real life recreation: 8/10, the street light is easy to recreate with metal and 
light bulbs 

Calculation: Considering the weight distribution and scores above, we have 
(10/10) * 0.5 = 0.5, 
(10/10) * 0.3 = 0.3, 
(10/10) * 0.15 = 0.15, 
(8/10) * 0.05 = 0.04. 
Adding the results up we have 0.5 + 0.3 + 0.15 + 0.04 = 0.99 

Final probability score: 0.99
\end{verbatim}
\end{framed}

\subsection{Investigating Open-Sourced Multimodal LLMs}
\label{appendixa3}
Further exploration into the LLaVA demo model \cite{liu2023improved} presents additional considerations. LLaVA faces a constraint with input token limits, necessitating a revision of the prompt outlined in Appendix \ref{appendixa1}. Additionally, LLaVA's inconsistent realism scores upon repeated queries signal a lack of reliability as a Judge, underscoring the need for targeted fine-tuning to bolster its evaluative precision.

\includegraphics[width=1\textwidth]{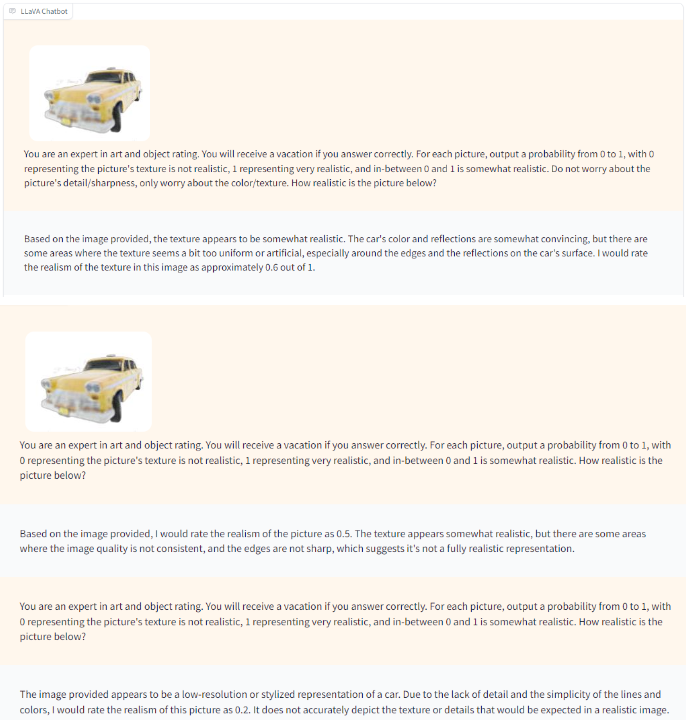}

\end{document}